\documentclass[11pt,oneside,a4paper]{article}
\usepackage[a4paper,left=2.5cm,right=2.5cm,top=2cm,bottom=2cm]{geometry}
\usepackage{amsfonts}
\usepackage{graphicx}

\begin{document}
\title{Data Augmentation with Diffusion Models for Colon Polyp Localization on the Low Data Regime: How much real data is enough?}
\author{Adrian Tormos\footnotemark[1], Blanca Llauradó\footnotemark[2], Fernando Núñez\footnotemark[2], Axel Romero\footnotemark[2],\\ Dario Garcia-Gasulla\footnotemark[1], Javier Béjar\footnotemark[2]}
\date{\small\footnotemark[1] Barcelona Supercomputing Center (BSC) \\
High Performance Articicial Intelligent group (HPAI)\\
\footnotemark[2] Technical University of Catalonia (UPC, BarcelonaTech)\\
Knowledge Engineering and Machine Learning Group (KEMLg)\\
Barcelona, Spain}
\maketitle

\begin{abstract}
The scarcity of data in medical domains hinders the performance of Deep Learning models. Data augmentation techniques can alleviate that problem, but they usually rely on functional transformations of the data that do not guarantee to preserve the original tasks. To approximate the distribution of the data using generative models is a way of reducing that problem and also to obtain new samples that resemble the original data. 
Denoising Diffusion models is a promising Deep Learning technique that can learn good approximations of different kinds of data like images, time series or tabular data. 

Automatic colonoscopy analysis and specifically Polyp localization in colonoscopy videos is a task that can assist clinical diagnosis and treatment. The annotation of video frames for training a deep learning model is a time consuming task and usually only small datasets can be obtained. The fine tuning of application models using a large dataset of generated data could be an alternative to improve their performance. We conduct a set of experiments training different diffusion models that can generate jointly colonoscopy images with localization annotations using a combination of existing open  datasets. The generated data is used on various transfer learning experiments in the task of polyp localization with a model based on YOLO v9 on the low data regime.

\end{abstract}

\section{Introduction}
The availability of large datasets for training deep learning models is an issue on many medical applications. On the one hand, to recruit a large number of patients and control individuals for recording the data is usually unfeasible for many domains; on the other hand, the annotation of the data is time consuming and also requires specialized knowledge.

Generative deep learning models present an opportunity to leverage the availability of smaller datasets for data augmentation. These models allow to learn the distribution of the datasets that can be then used for sampling from the data space, thus obtaining new samples that can be variations or combinations of the original data. As a result, richer datasets can be obtained so as to train more robust deep learning models. 

In the domain of colonoscopy analysis, during the last years several public colon polyp datasets have been made available~\cite{ali2023multi,duc2022BKAI,houwen2023computer,li2021colonoscopy,ma2021ldpolypvideo,misawa2021SUN,pogorelov2017kvasir,silva2014toward}. One of the characteristics of these datasets is their wide differences on characteristics, resolution and quality. However, their size generally ranges from the hundreds to the tens of thousands of images, which is a very small amount of data when compared to some text modelling~\cite{pile} and image description~\cite{schuhmann2022laion} datasets -- or even the amounts of data available for other medical tasks~\cite{ikezogwo2023quilt}. Given deep learning models' need for very large datasets -- usually orders of magnitude bigger than what is currently available -- it is interesting to compensate the lack of available data by obtaining generative models that combine this diversity to build a larger ensemble of data~\cite{du2023arsdm,zhang2022expanding}. These datasets can be used as pretraining data or as additional data for solving different tasks. 

In this paper we present experiments with a number of diffusion model training methods using heterogeneous datasets. We have applied various techniques for obtaining models that can generate data at a target resolution for a downstream task. The approaches are based on the use of upscaling using an autoencoder, finetuning a lower resolution model, training with a dataset with mixed resolutions, and the use of upscaled generated data for training jointly with real data at the target resolution. For testing the effectiveness of these methods we use different measures of data quality defined for generative models.

Additionally, we study the potential of exploiting synthetic data with transfer learning on a low-data availability downstream task, polyp localization. We perform a pre-training with our generated data before training the models with real data, and compare performance on the selected task when performing and skipping the pre-training stage. We also assess the amount of real data in which including the pre-training stage results in bigger gains.

\section{Related work}

The use of generative models as a source for training or pretraining data has been tested on different domains since its successes on the generation of realistic data in many domains. Recently, diffusion models have taken over the generative models area due to its higher generation quality and have been applied to obtain training data for different medical domains.

One of the first works that was published on training and pre-training models using generated data from a diffusion model~\cite{he2022synthetic} uses a text to image diffusion model pre-trained on ImageNet to generate new datasets. These are used as data augmentation on different image recognition tasks. The target was a set of diverse coarse and fine classification datasets. The models were built by fine-tuning a CLIP model using only generated data or a combination of generated and real data. They obtained good results compared with random initialization and ImageNet pre-training. Zhang et al.~\cite{zhang2022expanding} used diffusion and masked models for the generation of different datasets including medical data, such as pathMNIST, organMNIST and breastMNIST for several image classification tasks, testing different classifier architectures and comparing with usual image data augmentation techniques as CutMix~\cite{yun2019cutmix} or RandAugment~\cite{cubuk2020randaugment}.

Specifically in the polyp detection domain, Du et al.~\cite{du2023arsdm} use a diffusion model for generating data for polyp segmentation and localization. The diffusion model uses the mask associated to the images as conditioning for the generation. The usual diffusion loss is modified by adding a weighting based on the segmentation mask. An additional pretrained segmentation model is used for obtaining a mask from periodically generated data from the model during training. These data are used to define an IoU and a binary cross entropy loss that is used as additional feedback for the training. 

Macháček et al.~\cite{machacek2023maskconditioned} trained a diffusion model for generating segmentation masks that was then used for conditioning a diffusion model to generate polyp images. Different architectures for segmentation were trained with datasets composed by varying amounts of real and generated data, but focused on the small data regime. 

Finally, Pishva et al.~\cite{pishva2023repolyp} trained a FastGAN for generating segmentation masks.  These masks were used to train a diffusion model with unlabeled colon images cropped using the generated masks.  This model was after finetuned using cropped out polyps. A second diffusion model was trained using unlabeled colon images and finetuned with clean colon images. The images of cropped polyps generated with the first diffusion model were inpainted with the second model using the RePaint strategy \cite{lugmayr2022repaint} to obtain the final images. The generated data was used to train a U-Net based segmentation model with a combination of real and generated data.

\section{Diffusion models}

Denoising Diffusion Probabilistic models (DDPM) \cite{ho2020denoising} are a type of generative models that are based on connecting the real data distribution to the gaussian distribution. This is done through a process composed by a sequence of intermediate steps that are defined as a markov chain. This process is defined so it can be applied forward, so the original distribution is transformed into gaussian noise and also in reverse, so data can be obtained starting from gaussian noise. The constrain that guarantees these transformations is that the transitions between consecutive steps must be gaussian. This is true when the number of steps between the original distribution and the final gaussian noise is large enough (in the range of thousands).

The forward process can be defined as an exact computation given that is a combination of additive gaussian noises, allowing to perform efficient training by obtaining samples at different time steps. A neural network is trained using samples obtained from the forward process to predict the added noise between consecutive time steps. This allows to learn the corresponding reverse process. Sampling can be done by starting with gaussian noise, and iteratively predicting the noise to be eliminated from the current noisy sample that transform it into the next less noisy sample until a sample of the original distribution is obtained at the last step.

The formulation of DDPM uses the ELBO of the joint distribution of the forward and reverse processes distributions to define a loss for training. This loss is simplified considering constant the variance of the distributions and giving to all the denoising steps the same weight. This defines a simple loss based on the MSE between the noise added to a sample and the noise predicted by a neural network:
\[{\cal L} = \mathbb{E}_{t, x_t, \epsilon}[||\epsilon - \epsilon_\theta( x_t, t)||^2\]
where $\epsilon$ is the noise introduced at step $t$, $x_t$ is the noisy sample and $\epsilon_\theta$ is the neural network that predicts the noise from the noisy sample and the current time step. Equivalent formulations can be obtained for predicting directly $x_t$ instead of the noise or using a combination of both that can be used as alternatives for optimization.

Data generation in these models can be conditioned to different side information. For instance, a class, a semantic segmentation masks or a depth map among many others. This conditioning can be introduced in different ways in the denoising network to influence the sampling.

The network chosen for predicting the noise/sample is usually a U-Net with numbwer of parameters that scales with the size of the samples. In the case of images, this model could be very computationally costly for high resolutions. Latent Diffusion Models (LDM) \cite{rombach2022high} introduce a VAE or a VQGAN that transforms from the pixel space to a smaller latent space that is used for training the diffusion model. This reduces the computational cost of the training. In practice, a VAE/VQGAN pretrained with a very large dataset is used to avoid to have to train jointly both models.

\section{Datasets}
\label{datasets}

\paragraph{Generative model training} For the training of the generative models we selected four datasets that provided frames from colonoscopy videos with annotations (localization/mask): LDPolyp~\cite{ma2021ldpolypvideo}, SUN~\cite{misawa2021SUN}, PolypGEN~\cite{ali2023multi} and BKAI-IGH NeoPolyp-Small~\cite{duc2022BKAI}.

The LDPolyp dataset contains video frames of resolution $540\times 480$. It contains 40,266 frames where the training partition contains 24,789 frames. The training partition was selected for training the generative models. As the dataset is formed by videos, there are many consecutive frames that have very small differences, so they can be deemed duplicated samples. One problem with this circumstance is that diffusion models can end up memorizing samples when very similar or duplicated instances appear in the training dataset \cite{somepalli2023understanding}. To reduce this problem we decided to apply a deduplication strategy reducing the number of images that were to similar to each other.

The deduplication was computed using the features  obtained from the last layer before the classification layers of  VGG19 network pretrained with Imagenet. This transformed the images to vectors of 4096 dimensions. The distribution  of the euclidean distances to the first neighbor of each image in this embedding space was used to decide a threshold to build a neighborhood graph. This graph contained all the edges shorter that the threshold. All connected components of the graph were considered duplicated or quasi duplicated sets and for all the connected components with more than one image only one image was kept. This reduced the dataset to  17,360 unique images.

The SUN dataset contains video frames of resolution $1158\times 1008$. It contains 49,136 frames where the training partition has 16,497 frames selecting only the videos that contain polyps. We also selected the training partition for the training the generative models. In this case we did not deduplicated the dataset to maintain a reasonable dataset size, in consequence, this dataset has less diversity than the rest.

PolypGEN is a multicenter dataset that joints six different datasets with different resolutions but usually larger than $1200\times 1080$. From all the frames we selected the first five sources since the sixth one had lower image quality, with samples with different illuminations, blurriness and time stamps over the image. For the rest of the sources, only the part of the image corresponding to the colonoscopy was kept cropping other information like the position of the endoscope or time stamps. Since the frames did not correspond to sequential video frames deduplication was not needed. The final dataset contained 1,423 images.

The BKAI-IGH NeoPolyp-Small dataset contains video frames of resolution $1280\times 959$. It has 1,200 frames where the training partition has 1,000 frames. We also selected the training partition for the training the generative models. These frames are not sequential video frames, so deduplication was not applied either.

\paragraph{Transfer learning} 

On the transfer learning experiments, we re-use small random subsets of various different sizes of LDPolyp, SUN and PolypGEN to train polyp localization models and combine them with our synthetic data (see further details on Section~\ref{sec:transfer-learning}).

\paragraph{Evaluation}

Transfer learning experiments are 0-shot evaluated on KVASIR-SEG~\cite{jha2020kvasir}, ETIS-Larib~\cite{silva2014toward}, POLAR~\cite{houwen2023computer}, and KUMC~\cite{li2021colonoscopy}.

The KVASIR-SEG dataset contains 1,000 variable sized images, ranging from $332\times 487$ to $1920\times 1072$ pixels. ETIS-Larib contains 197 polyp images of resolution $1225\times 966$. The POLAR database is formed by a train (2,637 images) and a validation (730 images) split, both containing images of around $1000\times 1000$ pixels, of which we use the latter on our evaluations. We also only use the test split of the KUMC dataset, which contains 4,872 images of around $500\times 500$ pixels.

All datasets contain bounding box annotations on all images. On all datasets, the instances over $500\times 500$ pixels have all been resized to size $640\times 640$.

\section{Polyp generation}
\label{sec:polyp-generation}

As model for data generation we have used a conditional Latent Diffusion Model (LDM)~\cite{rombach2022high}. To reduce computational cost the pretrained VAE from the Stable Diffusion XL~\cite{podell2023sdxl} has been used to preprocess the data and obtain a dataset in latent space. This VAE has been trained on a very large dataset and is able to obtain very good reconstructions.

\begin{figure}[t]
\begin{center}
    \includegraphics[width=\textwidth]{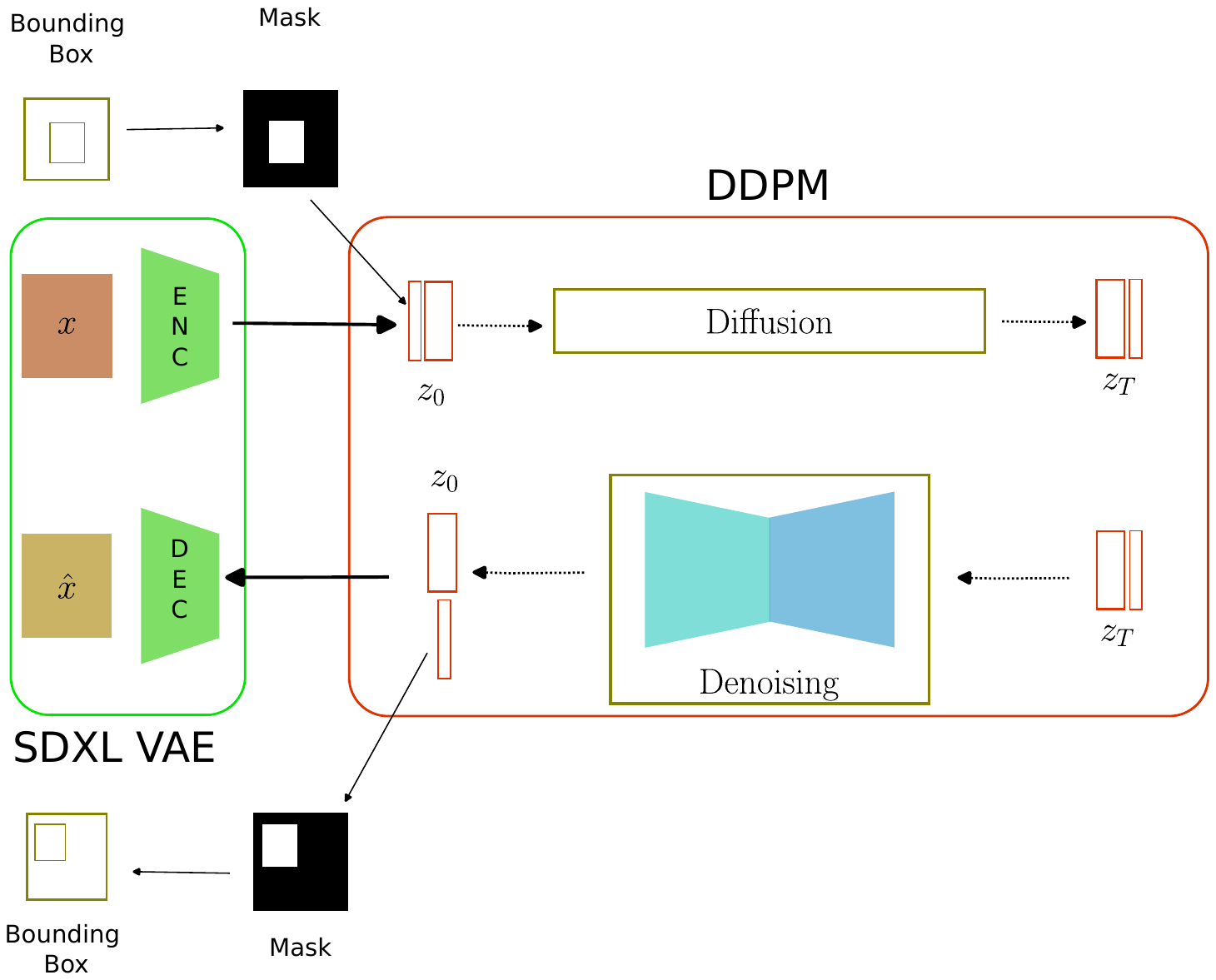}
\end{center}
\caption{\label{fig:diffusion} Conditioned Latent Diffusion. The polyp images are transformed using the SDXL VAE to a latent representation, bounding boxes are transformed to a binary masks, downscaled and appended as additional channels to the latent representation, diffusion is applied to train a denoising U-Net network that genrerates a latent image and a binary mask, from the binary mask bounding boxes are computed and the latent representation is decoded using the VAE.}
\end{figure}

The model has been conditioned to a binary square mask obtained from the polyp localization bounding boxes where available or using the segmentation map and computing the bounding box that contained the polyp mask. This square mask was added as an additional channel to the image avoiding the use of an additional embedding for conditioning. The masks were downscaled to the size of the spatial of latents ($32\times 32$) since the pretrained VAE we have used preserves the spatial information of the original image.

Using the mask as an additional channel fulfills a double purpose, the conditioning of the generation and the possibility of generating jointly the images and the masks when sampling from the model. This obtains the annotations for the images where the predicted mask can also be transformed into a bounding box for the localization model. See figure \ref{fig:diffusion} for a representation of the diffusion process.

As mentioned in section \ref{datasets}, the selected datasets have different resolutions, qualities and sizes. Since LDPolyp is the largest dataset the idea is to use it as the main source of the training, using the rest as complementary data. 

One of the goals of the experiments was to obtain a dataset with an intermediate resolution (not too small, but not too large), deciding for $640\times 640$ for the experiments. The smaller datasets have larger resolutions and LDPolyp has smaller resolution. This can be solved in different ways, with upscaling and downscaling being the simplest solutions. Usually downscaling is less problematic than upscaling, so we decided to downscale the SUN, PolypGEN and NeoPolyp datasets and to work with the LDPolyp dataset at its native resolution.

Since we are using a pretrained  VAE and a U-Net denoising network that are fully convolutional,  we took advantage of the possibility of using larger latent noise samples as input for the denoising network. Combined with the VAE decoder, this obtains upscaled samples proportional to the size of the latent spatial resolution. We observed that when the difference between the training resolution and the target resolution is not too large the VAE, is able to obtain good quality images at higher resolution.

For training diffusion models with all the datasets we planned the following experiments:

\begin{enumerate}
\item \textbf{VAE upscaling:} To obtain a center crop of all the images and downscaling all the images to the native resolution of LDPolyp ($480\times 480$), to train the LDM with the joint dataset and generate samples using $80\times 80$ latent gaussian noise to obtain the $640\times 640$ target resolution
\item \textbf{Finetuning:} To obtain a center crop of all the images, downscaling SUN, PolypGEN and NeoPolyp to $640\times 640$, maintaining LDPolyp to its native resolution, to train a LDM with LDPolyp and then finetune the model using the rest of the data at a lower learning rate, samples are generated  using $80\times 80$ latent gaussian noise 
\item \textbf{Alternate batch:} To obtain a center crop of all the images, downscaling SUN, PolypGEN and NeoPolyp to $640\times 640$, maintaining LDPolyp to its native resolution, and to train a LDM using an alternate batch training, where each $480$ resolution batch from LDPolyp is followed by a $640$ resolution batch of samples from the rest of the datasets
\item \textbf{Alternate epoch:} To obtain a center crop of all the images, downscaling SUN, PolypGEN and NeoPolyp to $640\times 640$, maintaining LDPolyp to its native resolution, and to train LDM using an alternate epoch training, where each $480$ resolution epoch is followed by a $640$ resolution epoch
\item \textbf{Mixed generated and real:} To obtain a center crop of all the images, downscaling SUN, PolypGEN and NeoPolyp to $640\times 640$, maintaining LDPolyp to its native resolution, and to train a first LDM using only LDPolyp. This model is then used to generate a large dataset (80,000) of images at  $640\times 640$ resolution using the VAE as upscaler and these data are joined with the other datasets to train a second LDM model at the target resolution
\end{enumerate}

The denoising network used in the Latent Diffusion model is the usual U-Net architecture with three encoder blocks, a middle block and three decoder blocks. Each block is composed by three layers of residual 2D convolutions adding self attention only in the last residual block of the encoder and the first residual block of the decoder. The training was performed using the DDPM algorithm predicting the noise of the sample using 1000 denoising steps.

The scheduler used for the noise generation was linear. Also, the zero final PSNR method was used for the scheduler during training to allow the generation of the full range of brightness in the images. It has been observed that schedulers do not end on a unit gaussian distribution on the last step, compromising the range of brightness of the images, this strategy corrects this problem, assuring that the last step correspond to full gaussian noise. 

 The training of all the models was done in the same way using adam as optimizer with a learning rate of $10^{-4}$ with cosine  schedule. The training used an EMA of the model. Augmentation was used for training using geometrical transformations (horizontal and vertical flip, transposition and rotation 90/180/270). Each model was trained using a single GPU (nvidia H100) for 48 hours.

After training, a dataset of $80,000$ samples was generated for each model using DDPM a sampler with 300 denoising steps. To test the quality of the dataset the Frechet Inception Distance (FID) and the Inception score (IS) using 4096 features were computed for all the datasets. Table \ref{tab:FID} shows that the dataset obtained with the model on the reduced resolution with the following upscaling with the VAE has the highest FID, followed by the mixed resolution training with alternating epoch training. For the IS these two models are also the best ones.

\begin{table}[t]
    \centering
    \begin{tabular}{c|c|c|c|c|}
        Experiment & FID & IS &Precision & Recall\\\hline
        VAE Upscale &31.75 &1.0676 &0.231&0.332\\ 
        Finetuning  &33.67 &1.0615&0.275&0.210\\ 
        Alternate Batch& 41.00&1.0658&0.184&0.303\\ 
        Alternate Epoch& 32.50& 1.0702&0.218&0.450\\ 
        Mixed& 64.54&1.0543&0.242&0.220\\ 
    \end{tabular}
    \caption{Quality measures for the generated datasets }
    \label{tab:FID}
\end{table}

When computing the FID of the real training data with the real test data a value of 17.41 is obtained. One possible reason for this difference with the generated data is the use of augmentations during the training of the models, that probably increases the diversity of the generated images and obtains out of distribution samples. It also could be due to the relative small size of the training data.

The IS of the train dataset is 1.0677, that is very similar to the IS of all the generated datasets. This means that this score is not a good measure for evaluating the quality of the data for this particular dataset. This is not surprising since all images are very homogeneous and are not related to any of the classes on the Imagenet dataset that is the base of the Inception Score.

Additionally, we computed the precision and recall measures \cite{kynkaanniemi2019improved} that are an approximation of their supervised counterparts. These measures use features from the InceptionV3 network for computing the intersection between the distribution of the real images and the generated images and vice-versa using neighborhood graphs. The results (see table  \ref{tab:FID}) show that the best recall corresponds to the alternate batch model and the best precision to the finetuning model. The results are relatively low probably also because of the augmentations that were used for the training of the models.

\section{Transfer learning in a low-data availability setting}
\label{sec:transfer-learning}

We study a transfer learning setting in which we have low amounts of data available.  This is a plausible situation, given that capturing a large number of videos from colonoscopy and curating a dataset with a specific set of characteristics suitable to a model (\textit{e.g.} images higher than a specific resolution) is expensive and time consuming. In which case, we may resort to the use generative models trained on heterogeneous data to create synthetic images to supplement our available data.

We use a YOLOv9~\cite{wang2024yolov9learningwantlearn} localization network, pre-trained with ImageNet. We train the model with different combinations of real data and the datasets obtained with generative models in Section~\ref{sec:polyp-generation}. In our setting, our low-availability real data are subsets (10, 25, 50, 100, 250, 500 and 1000 images) of a mixture of the SUN, LDPolyp and PolypGen datasets and our generated data are the five datasets obtained in Section~\ref{sec:polyp-generation}. 

We evaluate the performance of the YOLO model in two modalities of training: (A) when trained just with our low-availability real data, and (B) when first trained with 80,000 synthetic images and then fine-tuned with our low-availability real data.

In the trainings using the synthetic data, the networks were trained for 13 epochs (batch size 32) using the SGD optimizer with momentum and weight decay. In the trainings with the real data, the networks were trained for $\frac{10000}{bs\cdot n}$ epochs, where $bs=32$ (batch size) and $n$ is the size of the dataset, also using the SGD optimizer with momentum and weight decay. Each experiment has been performed with 3 different seeds. Real images are scaled to 640$\times$640 pixels.

The resulting networks were evaluated 0-shot in the test split of four polyp detection datasets: KVASIR~\cite{pogorelov2017kvasir}, ETIS-Larib~\cite{silva2014toward}, POLAR~\cite{houwen2023computer} and  KUMC~\cite{li2021colonoscopy}.

\begin{figure}[t]
\begin{center}
    \includegraphics[width=\textwidth]{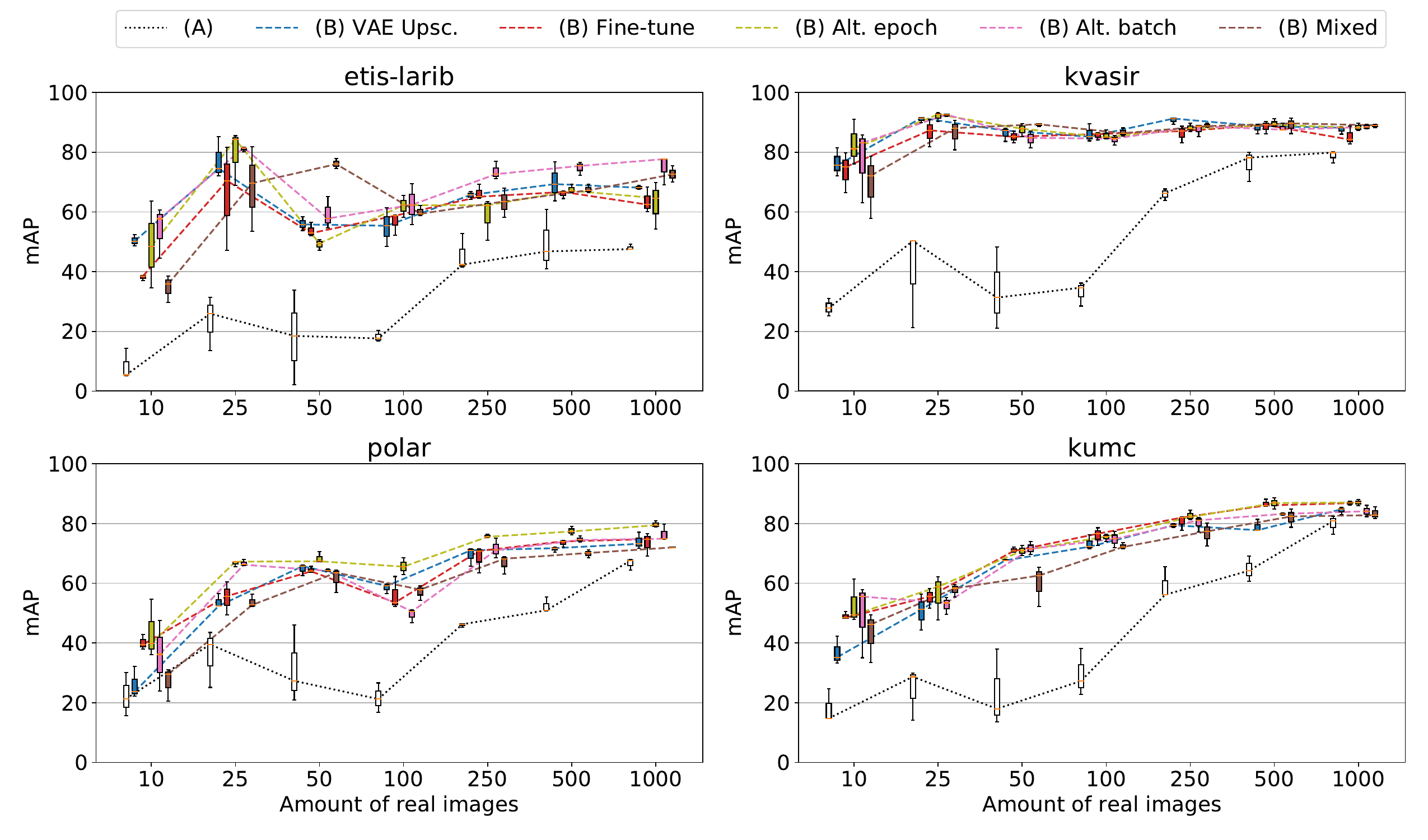}
\end{center}
\caption{\label{fig:results} Mean Average Precision (mAP) of the trained YOLOv9 networks on the four evaluation datasets. The dotted series correspond to networks trained with training modality A. The rest of the series correspond to modality B, each of the differently-coloured series using one of the five presented synthetic datasets.}
\end{figure}

In Figure~\ref{fig:results} we observe that the performance of the trained networks generally increases with the size of the dataset. The performance of networks trained with modality B plateaus in the KVASIR and ETIS-Larib datasets, while it keeps increasing on the remaining two datasets. As opposed to these networks, which already show performance improvements on the smallest dataset sizes, networks trained with modality A show constant, poor performance until the 250-image mark, from which point onward they show jumps of 25 to 40 mAP percentage points in all benchmarks. Nonetheless, networks trained with A only match the performance of the other networks in a selection of trainings with either very little data (10 images) or a relatively large dataset (1000 images).

It can be seen on all four evaluation datasets that training modality B is especially effective up to the 100-image mark, achieving around 2 to 4 times the performance of training modality A when training with 50 and 100 images. On very small trainings (10 images), training modality B achieves similar or better performance gain on 3 out of 4 datasets. On 250-image or higher trainings, modality B still performs 1.1 to 1.5 times better than modality A.

Note the variation among results on the different datasets. It may be explained by their inherent complexity and difficulty. For instance, networks trained with with generated data and as little as 25 real images are enough to perform well on KVASIR. As dataset size increases on POLAR and KUMC, the performance gain is steady but the advantage obtained from the pre-training diminishes. The networks' behaviour on ETIS-Larib seems to be more irregular, with networks trained with only 25 images achieving slightly better performance than with larger amounts of data. 

As for the different types of generative models used for obtaining the pre-training dataset, they all generally seem to follow a narrow band of performance gain for all the experiments that is especially evident on the KVASIR and KUMC datasets. None of the five datasets consistently results in comparatively better or worse performance across evaluations. For most of the experiments, especially training on 50 or more images, the variation of performance for different initializations is small, indicating a robustness of the pretrained models. Surprisingly, the results of the different generative models are not linked to the quality measures. It seems that even when the data generated by the models are not close to the distribution of the original datasets, the synthetic data that is generated is helpful for pre-training.

\section{Conclusions}

The results of the experiments show the benefit of using synthetic data as a pre-training dataset when the available data for training a model is not enough for obtaining a good performance in the domain of polyp localization. 

As it could be expected, the gain obtained with the pre-training phase is larger for smaller datasets, and almost converges to the results when skipping it as the dataset gets larger, so the cost-benefit trade off between gathering more data or using synthetic data has to be taken in account in order to decide to use it or not.

The way the generative model is trained seems not to have a huge impact on the quality of the pretrained network. Our intuition is that the use of many diverse datasets for training the generative models that mixes quality and variety is a larger factor on the success of this pre-training strategy. This means that probably the simplest technique for training the generative model is enough for obtaining the synthetic data.

In our opinion, to have generative models trained on a combination of multiple datasets has several advantages. The first one is the capability of the models of generating pretraining datasets of any size that cover a large variety of samples. The second one is the possibility to fine tune the generative models to the specific dataset that will be used to obtain the predictive model using for instance LoRA \cite{hu2021lora} and to obtain a pretraining dataset more similar to the target dataset. This is one of our future work, as our intuition is that a fine tuned generative model could reduce more the need of data collection.

The availability of generative models for different types of medical data can helpful for the development of specialized models with reduced data gathering. It is a fact for instance, that the equipment and preprocess used for the collecting of data in different hospitals can be very different and that can be a barrier for the development of models adjusted to their particular needs. To use data from generative models can open the possibility for models that suit these needs.

\section{Acknowledgments}

 This research has been funded by the Artificial Intelligence for Healthy Aging (AI4HA, MIA.2021. M02.0007) project from the Programa Misiones de I+D en Inteligencia Artificial 2021 and by the European Union-NextGenerationEU, Ministry of Universities and Recovery, Transformation and Resilience Plan, through a call from Universitat Politècnica de Catalunya and Barcelona Supercomputing Center (Grant Ref. 2021UPC-MS-67461/2021BSC-MS-67461).
\bibliographystyle{plain}
\bibliography{paper}
\end{document}